%% file: AnonymousSubmission2027.tex
\documentclass[letterpaper]{article} 
\usepackage{aaai2027} 
\usepackage[hyphens]{url} 
\usepackage{graphicx} 
\usepackage{natbib} 
\usepackage{caption} 
\usepackage{amsmath,amsfonts}
\usepackage{algorithm}
\usepackage{algorithmic}
\usepackage{array}
\usepackage{booktabs}
\usepackage{colortbl}
\usepackage{multirow}
\usepackage{pifont}
\usepackage{tabularx}
\usepackage{xcolor}

\newcommand{\vcenteredbox}[1]{%
  \begingroup
  \setbox0=\hbox{#1}%
  \parbox{\wd0}{\box0}%
  \endgroup
}

\title{Geospatial-Prior Guidance for 3D Semantic Scene Completion}
\author{Meng Wang, Shougao Zhang, Wenzhe He, Ruihui Li, Nan Hu, Zhuo Tang, Kenli Li}
\affiliations{
  College of Computer Science and Electronic Engineering, Hunan University, Hunan, China\\
    \{willem, zhangshougao, hewenzhe, liruihui, hunan5, ztang, lkl\}@hnu.edu.cn
}

\begin{document}

\maketitle

\input{sec/0_abstract}

\begin{figure*}[t]
  \centering
  \includegraphics[width=\textwidth]{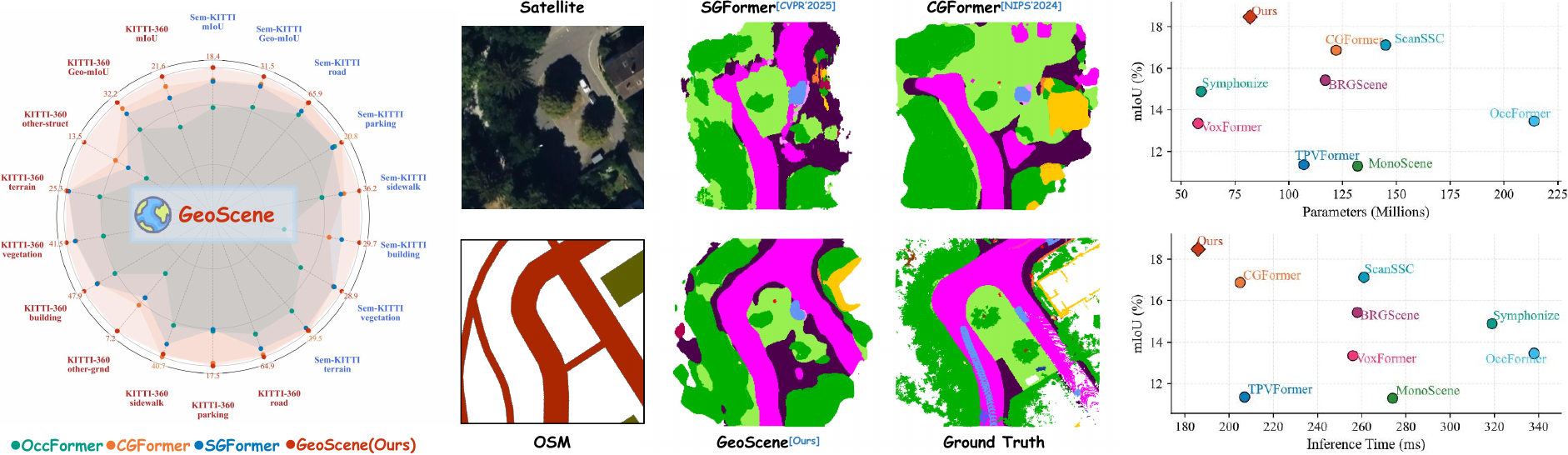}
  \caption{GeoScene integrates geospatial priors to guide 3D semantic scene completion. Compared with previous SSC methods, GeoScene produces geometrically complete and semantically consistent reconstructions by leveraging both observation and geospatial priors.}
  \label{fig:figure1}
\end{figure*}

\input{sec/1_intro}
\input{sec/2_related}
\input{sec/3_method}
\input{sec/4_exp}

\bibliography{sample-base}

\clearpage
\appendix
\input{sec/X_suppl}

\end{document}

%% file: sec/0_abstract.tex
\begin{abstract}
Inferring complete 3D geometry and semantics from onboard images remains challenging because occlusions and restricted fields of view leave large scene regions underconstrained.
Although satellite imagery provides wide-area context, appearance cues alone offer limited structural guidance and may be unreliable because of spatial or temporal discrepancies.
We present GeoScene, a geospatially guided framework that jointly uses satellite imagery and structured OpenStreetMap cues as soft priors for 3D semantic scene completion.
GeoScene learns complementary voxel-wise reliability weights for onboard observations and geospatial guidance, and uses them to control feature refinement in observed and unobserved regions.
This design preserves local visual evidence while exploiting large-scale road and building structure beyond onboard visibility.
Experiments on SemanticKITTI and SSCBench-KITTI-360 demonstrate that GeoScene consistently improves both geometric and semantic completion under the geospatial-prior-assisted setting, with the most pronounced benefits for large-scale static and geospatially structured classes.

\end{abstract}

%% file: sec/1_intro.tex
\section{Introduction}
\label{sec:intro}

Understanding the complete 3D geometry and semantics of complex outdoor environments is a long-standing challenge in autonomous perception.
Reasoning beyond the visible field of view is particularly important for motion planning, autonomous navigation, and large-scale scene understanding, where safe decisions depend on both observed and unobserved regions.
To support such reasoning, 3D semantic scene completion (SSC)~\cite{song2017semantic,roldao2020lmscnet,WSSIC-Net} jointly predicts volumetric occupancy and semantic labels from partial 3D observations, thereby recovering a holistic representation of the surrounding environment.

Recent advances have shown that dense 3D semantic representations can be inferred directly from monocular RGB images~\cite{cao2022monoscene,zhang2023occformer,huang2023tri,jiang2024symphonize,li2023voxformer,li2024htcl,mei2024sgn,wang2024HASSC}, reducing dependence on expensive LiDAR sensors.
Beyond voxel- and transformer-based context aggregation, object-centric feature learning explicitly organises representations around scene entities to strengthen local geometry and semantic reasoning~\cite{wang2026objectcentric}.
Despite this progress, camera-based SSC remains fundamentally constrained by limited fields of view, occlusion, and viewpoint-dependent ambiguity.
Regions behind buildings, outside the camera frustum, or heavily occluded by foreground objects receive little direct evidence, often resulting in incomplete geometry and unstable semantic predictions.
Because these methods derive their context primarily from onboard observations, stronger local or object-level reasoning alone cannot supply large-scale structural constraints beyond the sensing range.

Recent studies~\cite{Guo_2025_sgformer,chen2025sa} have therefore explored overhead or satellite imagery as complementary context for 3D perception.
Although these methods demonstrate the value of overhead appearance cues, the joint use of satellite imagery and structured map-level semantics remains less explored.
Structured sources such as OpenStreetMap (OSM) encode human-interpretable layouts, including road networks and building footprints, and thus complement the visual cues provided by satellite imagery.
This observation motivates a fundamental question:
\textbf{\textit{Can we exploit geospatial priors to guide 3D semantic scene completion beyond onboard visibility?}} 

Geospatial information provides a form of context that is complementary to onboard visual evidence.
Whereas onboard cameras capture local and viewpoint-dependent observations, satellite imagery and map abstractions describe wide-area, topology-aware structures that are comparatively stable across viewpoints and short-term appearance changes.
These priors are especially informative for static elements such as roads, buildings, and terrain---structures that are difficult to infer from limited views but essential for globally coherent completion.
When integrated appropriately, they can provide soft spatial and semantic guidance for occluded or unobserved regions.

Directly injecting geospatial information into SSC is, however, non-trivial.
Such priors may be incomplete, noisy, spatially misaligned, or temporally outdated, and their reliability can vary across locations and semantic classes.
Treating them as hard constraints may therefore propagate erroneous structure, particularly for dynamic objects and fine-grained categories.
Effective integration must instead balance onboard evidence and geospatial guidance according to their local reliability.

In this work, we propose \textit{\textbf{GeoScene}}, \textbf{Geo}spatial-Prior Guidance for 3D Semantic \textbf{Scene} Completion, a framework that uses external geospatial cues as soft auxiliary priors rather than ground-truth supervision.
GeoScene jointly integrates satellite imagery and structured OSM cues to improve the reconstruction of large-scale static structures beyond onboard visibility.
At the core of GeoScene is a Dual-Priors Weighted Classifier, which learns two complementary voxel-wise soft reliability weights:
(i) an observation weight capturing the reliability of image-derived features, and
(ii) a geospatial weight controlling the contribution of external geospatial guidance.
These weights modulate feature fusion within a Weights-Guided Voxel Refiner, adaptively combining local observations with global structural cues while suppressing unreliable guidance.
We evaluate GeoScene on two challenging SSC benchmarks, SemanticKITTI~\cite{behley2019semantickitti} and SSCBench-KITTI-360~\cite{Liao2022kitti360,li2023sscbench}.
Extensive experiments demonstrate consistent improvements in the geospatial-prior-assisted setting, particularly for large-scale static and geospatially structured classes.
Our main contributions are summarized as follows:
\begin{itemize}
\item We introduce GeoScene, which jointly incorporates satellite imagery and structured OSM cues as soft priors for completing large-scale static structures beyond onboard visibility.
\item We propose a Dual-Priors Weighted Classifier that learns complementary observation and geospatial soft reliability weights, together with a Weights-Guided Voxel Refiner for adaptive voxel-level fusion.
\item Experiments on {SemanticKITTI} and {SSCBench-KITTI-360} demonstrate state-of-the-art performance under the geospatial-prior-assisted setting, with the largest benefits concentrated on static, geospatially structured classes.
\end{itemize}

%% file: sec/2_related.tex
\section{Related Work}
\label{sec:related}
\begin{figure*}[t]
\centering
  \includegraphics[width=\textwidth]{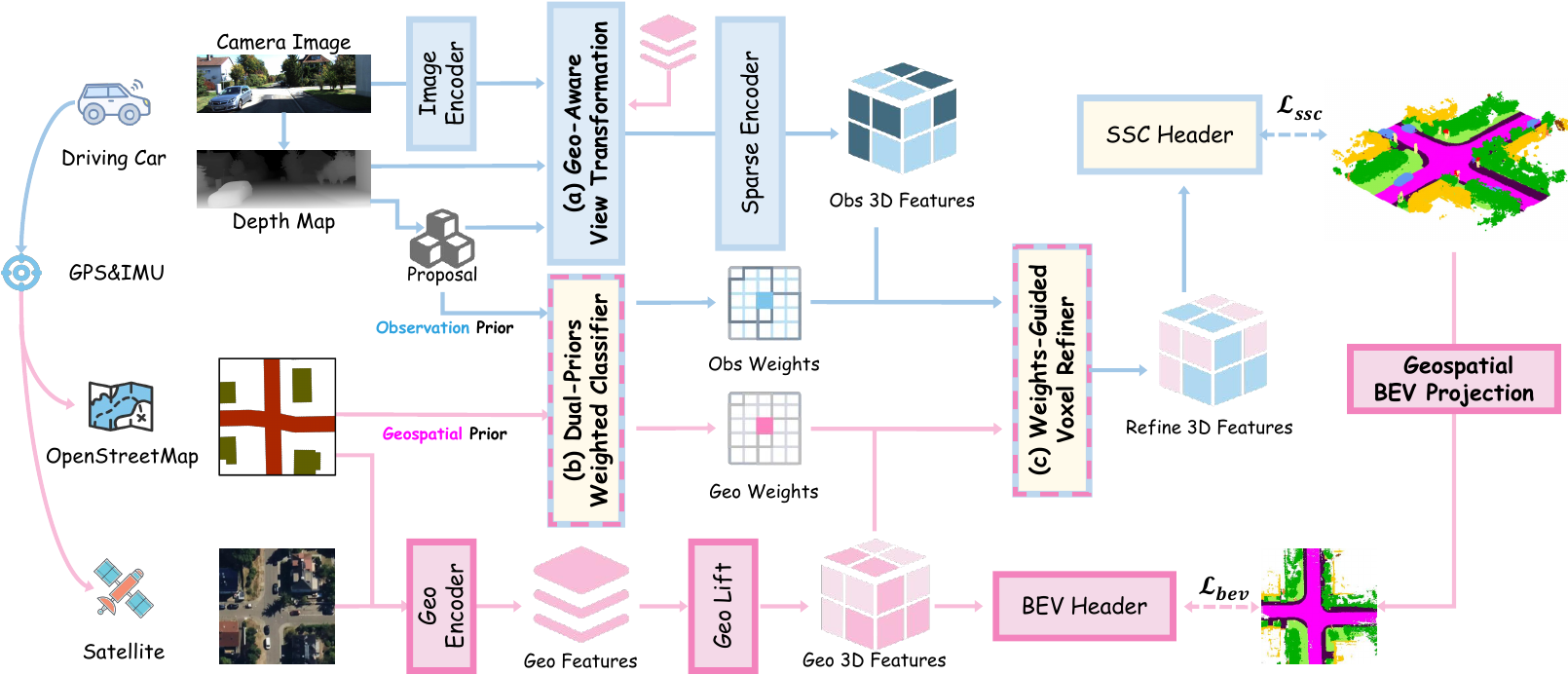}
  \caption{The GeoScene framework for geospatially guided 3D semantic scene completion.}
  \label{fig:figure2}
\end{figure*}

\noindent\textbf{3D Semantic Scene Completion.}
Semantic scene completion (SSC) reconstructs dense 3D occupancy and semantics from partial observations. Monocular methods have progressed from direct volumetric prediction~\cite{cao2022monoscene} to tri-plane and voxel-transformer representations~\cite{huang2023tri,zhang2023occformer,li2023voxformer,wei2023surroundocc}. Subsequent studies improved feature aggregation through cross-attention, instance-level reasoning, multi-view reconstruction, and self-distillation~\cite{zheng2024monoocc,jiang2024symphonize,wang2024h2gformer,wang2024HASSC}. Object-centric feature learning further organises voxel representations around scene entities to improve instance-aware geometric and semantic reasoning~\cite{wang2026objectcentric}. Stereo, depth, motion, and semantic priors also reduce geometric ambiguity~\cite{li2023stereoscene,wang2025mixssc,CGFormer,wang2025vlscene,Bae_2025_scanssc,EGS}. Nevertheless, these representations remain derived primarily from onboard observations and do not directly encode large-scale road--building structure outside the visible region.

\noindent\textbf{Geospatial Perception.}
Geospatial perception complements ground-level observations with wide-area structural context. Early work established cross-view correspondences for retrieval and localisation~\cite{hu2018cvm}, followed by pose-aware methods for finer geometric alignment~\cite{shi2022accurate,xia2023convolutional,wang2024fine}. BEV-based approaches subsequently supported dense satellite--street reasoning for segmentation and map reconstruction~\cite{ye2024cross,ye2024sg,sarlin2024snap}. Recent occupancy methods incorporated satellite features into 3D prediction~\cite{chen2025sa,Guo_2025_sgformer}, but satellite appearance alone may provide uncertain structural guidance. GeoScene instead combines satellite imagery with structured OSM priors and adaptively weights their contributions during voxel refinement.

%% file: sec/3_method.tex
\section{Method}
\label{sec:method}
\noindent\textbf{Problem Setup.}
Given an onboard RGB image $I$, a spatially corresponding satellite image $I_\mathrm{sat}$, and a structured semantic map $S_\mathrm{osm}$ derived from OpenStreetMap, GeoScene aims to recover the complete 3D geometry and semantics of the surrounding environment.
We discretize the target space into a voxel grid of size $N_X\times N_Y\times N_Z$ and define the label set as $\mathcal{C}=\{C_0,C_1,\ldots,C_M\}$, where $C_0$ denotes empty space and $C_1,\ldots,C_M$ denote the $M$ semantic classes.
The ground-truth scene is represented by $\hat{Y}\in\mathcal{C}^{N_X\times N_Y\times N_Z}$.
Our objective is to learn a parameterized mapping $Y=f_\theta(I,I_\mathrm{sat},S_\mathrm{osm})$, 
where $Y$ contains the predicted class scores for all voxels.
The final semantic occupancy grid is obtained by selecting the highest-scoring label at each voxel, and the model parameters $\theta$ are optimized to match the ground-truth grid $\hat{Y}$.

\noindent\textbf{Overview.}
As illustrated in Figure~\ref{fig:figure2}, the overall workflow of GeoScene is as follows.  
Camera images are first fed into an image feature extractor~\cite{wang2023repvit,lin2017fpn} to obtain 2D observation features $F_\mathrm{obs}$.  
Simultaneously, satellite imagery and structured semantic masks are processed by a geospatial encoder~\cite{he2016resnet,lin2017fpn} to generate 2D geospatial features $F_\mathrm{geo}$.  
An off-the-shelf depth estimator~\cite{shamsafar2022mobilestereonet,bhat2021adabins} predicts the depth map $D$ from driving images and identifies visible surface voxels, which serve as query proposals $\mathcal{P}$~\cite{li2023voxformer}.  
Subsequently, $F_\mathrm{obs}$, $F_\mathrm{geo}$, $D$, and $\mathcal{P}$ are fused and transformed into 3D voxel features $V_\mathrm{obs}$ via a Geo-Aware View Transformation module.  
The resulting features are processed by a Sparse Encoder~\cite{wang2025vlscene} to efficiently handle non-empty voxels.  
After geometric alignment, the 2D geospatial features are lifted into BEV space and expanded into 3D to obtain geospatial voxel features $V_\mathrm{geo}$.
Next, the query proposals $\mathcal{P}$ and structured prior map $S_{\mathrm{osm}}$ are passed into the Dual-Priors Weighted Classifier, which estimates two complementary soft reliability fields, $W_\mathrm{obs}$ and $W_\mathrm{geo}$, for onboard observations and geospatial guidance, respectively.  
These weights, together with voxel features from both views, are integrated by the Weights-Guided Voxel Refiner, which performs adaptive feature fusion and refinement.  
Finally, the refined features $V_\mathrm{fine}$ are upsampled and linearly projected to produce dense semantic voxel predictions $Y$. A detailed illustration of the three modules is provided in Figure~\ref{fig:detailed_modules} in the appendix.

\subsection{View Transformation}
\label{sec:GAVT}
Building on existing view-transformation methods~\cite{li2023stereoscene,CGFormer}, we incorporate geospatial cues to construct an observation voxel volume $V_\mathrm{obs}\in\mathbb{R}^{X\times Y\times Z\times C}$, where $X$, $Y$, and $Z$ are the spatial dimensions and $C$ is the feature dimension. Given the observation features $F_\mathrm{obs}$, geospatial features $F_\mathrm{geo}$, depth map $D$, and query proposals $\mathcal{P}$, we first estimate a depth distribution following LSS~\cite{philion2020lift}. The depth distribution is combined with $F_\mathrm{obs}$ to construct 3D features $F_\mathrm{3d}$, which are projected onto the voxel grid to obtain $V_\mathrm{obs}$. We then use $\mathcal{P}$ as queries in deformable cross-attention~\cite{zhu2020deforDETR} over $F_\mathrm{geo}$, allowing $V_\mathrm{obs}$ to incorporate geospatial context.
Finally, $V_\mathrm{obs}$ is processed by a Sparse Encoder~\cite{wang2025vlscene} to efficiently encode non-empty voxels.

\subsection{Dual-Priors Weighted Classifier}
\label{sec:DPWC}
The Dual-Priors Weighted Classifier estimates two complementary volumetric weight fields, namely an observation weight $W_\mathrm{obs}$ and a geospatial weight $W_\mathrm{geo}$, from the voxel proposals $\mathcal{P}\in\mathbb{R}^{X\times Y\times Z}$ and the 2D semantic prior $S_\mathrm{osm}$.
The observation weight characterizes the strength of onboard evidence at each voxel, whereas the geospatial weight controls the contribution of structured spatial cues. Together, they enable spatially adaptive voxel-level fusion.

\paragraph{Observation Weight}
The observation branch derives voxel-wise reliability from the local spatial cues encoded in $\mathcal{P}$.
A lightweight 3D CNN predicts the corresponding soft weights:
\begin{equation}
W_{\mathrm{obs}} = \sigma(\phi(\mathcal{P})),
\end{equation}
where $\phi(\cdot)$ denotes a convolutional projection and $\sigma(\cdot)$ is the sigmoid function.
The resulting map reflects the relative strength of visible geometric evidence at each voxel.

\paragraph{Geospatial Weight}
The geospatial branch combines local occupancy density with structured priors to estimate soft weights for unobserved or weakly observed regions.
The local density is first computed as
\begin{equation}
\mathrm{Density} = \mathrm{AvgConv3D}(\mathcal{P}).
\end{equation}
The OSM map $S_\mathrm{osm}$ is encoded and lifted into 3D to provide static cues such as road and building layouts.
Let $E_\mathrm{osm}$ denote the lifted OSM features. The density, $E_\mathrm{osm}$, and $W_\mathrm{obs}$ are fused by a learnable projection $\psi(\cdot)$ to predict
\begin{equation}
U_{\mathrm{base}} = \sigma\!\left(\psi\!\left([\mathrm{Density},E_{\mathrm{osm}},W_{\mathrm{obs}}]\right)\right).
\end{equation}
An unobserved-region factor $U_\mathrm{factor}=1-W_\mathrm{obs}$ suppresses redundant guidance in well-observed regions, while class-aware scaling emphasizes geospatially informative categories:
\begin{equation}
\widetilde{W}_{\mathrm{geo}} = U_{\mathrm{factor}} \odot U_{\mathrm{base}} \odot (r\,M_{\mathrm{rb}} + o\,M_{\mathrm{oth}}),
\end{equation}
where $M_\mathrm{rb}$ is the binary mask for road and building cells derived from $S_\mathrm{osm}$, $M_\mathrm{oth}=\mathbf{1}-M_\mathrm{rb}$, and $r$ and $o$ are the corresponding scaling coefficients.
The preliminary weight $\widetilde{W}_\mathrm{geo}$ therefore emphasizes structurally plausible regions that receive limited onboard evidence.

\paragraph{Complementary Alignment}
To balance the two weight fields, we normalize and align them within their respective spatial domains.
We first define the binary masks
\begin{equation}
M_{\mathrm{obs}} = \mathbb{I}[\mathcal{P}>0], \qquad
M_{\mathrm{geo}} = \mathbf{1} - M_{\mathrm{obs}}.
\end{equation}
The alignment operation (i) matches the peak magnitudes of $W_\mathrm{obs}$ within $M_\mathrm{obs}$ and $\widetilde{W}_\mathrm{geo}$ within $M_\mathrm{geo}$, (ii) adjusts their joint statistics toward a target mean, and (iii) clamps the resulting values to $[0,1]$:
\begin{equation}
(W_{\mathrm{obs}},\,W_{\mathrm{geo}}) \leftarrow 
\mathrm{Align}(W_{\mathrm{obs}},\,\widetilde{W}_{\mathrm{geo}};\,M_{\mathrm{obs}},\,M_{\mathrm{geo}}).
\end{equation}
This alignment promotes a complementary allocation of the two weights: onboard evidence dominates well-observed voxels, while geospatial guidance supports unobserved or occluded regions.

Thus, $W_{\mathrm{obs}}$ preserves reliable onboard evidence, while $W_{\mathrm{geo}}$ acts as a soft gate for satellite and OSM cues in weakly observed regions.
These learned weights represent relative reliability rather than calibrated probabilities, allowing the subsequent voxel refiner to reduce the influence of geospatial guidance that conflicts with local observations.

\begin{figure*}[t]
\centering
  \includegraphics[width=\textwidth]{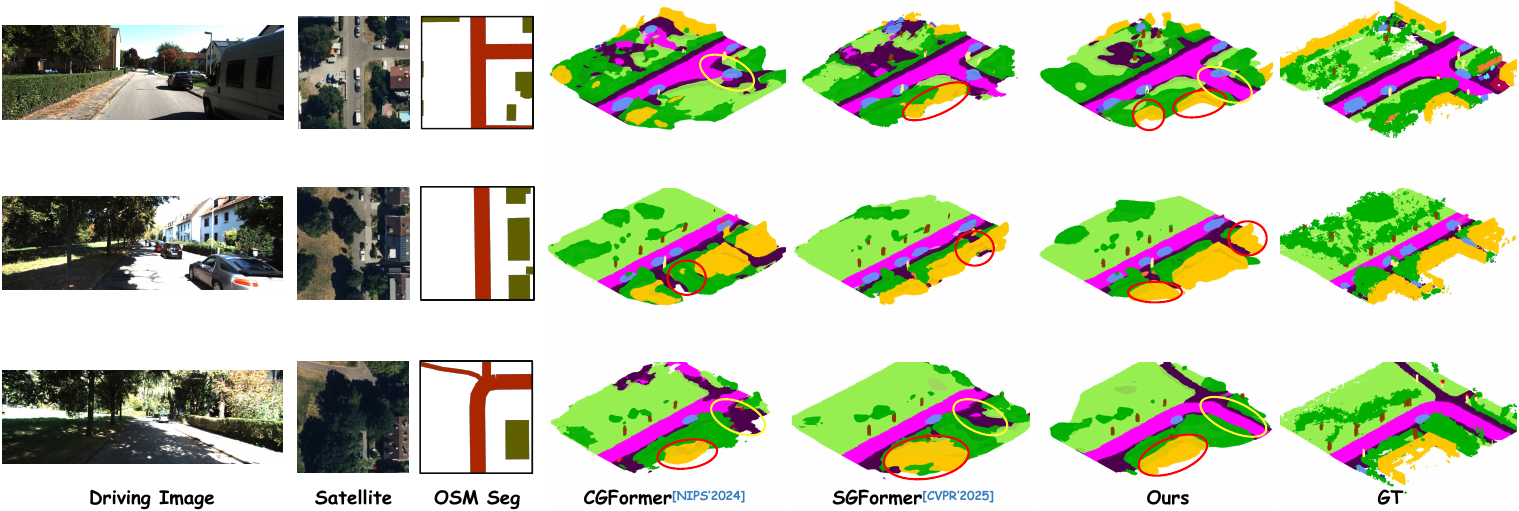}
  \caption{Qualitative comparisons on the SemanticKITTI validation set.}
  \label{fig:figure4}
\end{figure*}

\subsection{Weights-Guided Voxel Refiner}
\label{sec:WGVR}
The Weights-Guided Voxel Refiner uses the two complementary soft-weight fields to adaptively refine voxel features.
It contains an observation-oriented branch that preserves onboard evidence and a geospatial-oriented branch that introduces structural context into weakly observed or occluded regions.
Their outputs are combined through weight-guided fusion and subsequently processed by a global 3D harmonization stage to improve spatial and semantic consistency.

The satellite feature map is first broadcast along the vertical ($Z$) dimension to form the geospatial voxel representation $V_{\mathrm{geo}}$.
The refiner then receives $V_{\mathrm{obs}}$, $V_{\mathrm{geo}}$, and their corresponding weights $W_{\mathrm{obs}}$ and $W_{\mathrm{geo}}$.

The observation-oriented branch processes $V_{\mathrm{obs}}$ using two stacked $3{\times}3{\times}3$ convolutional blocks, denoted by $\operatorname{P}_{\mathrm{obs}}(\cdot)$:
\begin{equation}
\overline{V}_{\mathrm{obs}} = \operatorname{P}_{\mathrm{obs}}(V_{\mathrm{obs}}),
\end{equation}
where $\overline{V}_{\mathrm{obs}}$ preserves fine-grained geometric evidence from the street view.

In parallel, the geospatial-oriented branch aggregates the street-view and satellite features.
It employs $5{\times}5{\times}5$ and $3{\times}3{\times}3$ convolutional blocks, collectively denoted by $\operatorname{P}_{\mathrm{geo}}(\cdot)$, to capture broader structural context:
\begin{equation}
\overline{V}_{\mathrm{geo}} = \operatorname{P}_{\mathrm{geo}}\!\big([V_{\mathrm{geo}},\,V_{\mathrm{obs}}]\big),
\end{equation}
where $\overline{V}_{\mathrm{geo}}$ provides geospatially enriched representations for improving spatial continuity in unobserved regions.

The two branches are then combined according to their learned soft reliability weights:
\begin{equation}
V_{\mathrm{w}} = W_{\mathrm{obs}} \odot \overline{V}_{\mathrm{obs}} + W_{\mathrm{geo}} \odot \overline{V}_{\mathrm{geo}},
\end{equation}
where $\odot$ denotes element-wise multiplication. A lightweight 3D fusion block further refines $V_{\mathrm{w}}$ to harmonize the two feature sources.

To reduce scale imbalance between the two weights, we additionally compute a normalized fusion:
\begin{equation}
\begin{aligned}
W_{\mathrm{sum}} &= W_{\mathrm{obs}} + W_{\mathrm{geo}},\\
\overline{W}_{\mathrm{obs}} &= \frac{W_{\mathrm{obs}}}{W_{\mathrm{sum}}}, \qquad
\overline{W}_{\mathrm{geo}} = \frac{W_{\mathrm{geo}}}{W_{\mathrm{sum}}}.
\end{aligned}
\end{equation}
The normalized blend is
\begin{equation}
V_{\mathrm{norm}} = \overline{W}_{\mathrm{obs}} \odot V_{\mathrm{obs}} + \overline{W}_{\mathrm{geo}} \odot V_{\mathrm{geo}}.
\end{equation}

Finally, the two fusion outputs are combined using a fixed coefficient $\alpha$, followed by a global 3D harmonizer $t(\cdot)$ that promotes spatial and semantic consistency:
\begin{equation}
V_{\mathrm{fine}} = t(\alpha\,V_{\mathrm{w}} + (1-\alpha)\,V_{\mathrm{norm}}).
\end{equation}

\subsection{Geospatial BEV Projection}
To provide stable, geospatially consistent supervision, we derive a bird's-eye-view (BEV) semantic map from the 3D occupancy labels $\hat{Y}\in\mathcal{C}^{X\times Y\times Z}$.
Because static structures provide more reliable overhead cues than transient objects, we suppress dynamic categories and emphasize geospatially stable geometry during projection.
Specifically, the BEV target $\hat{\operatorname{B}}\in\mathcal{C}^{X\times Y}$ is constructed by scanning each vertical column $(x,y)$ within a valid height range and selecting the static class with the highest priority.
We use a dataset-specific order comprising (i) tall structures, including buildings, vegetation, and trunks; (ii) mid-level elements, including fences, poles, and signs; and (iii) ground-level classes, including roads, sidewalks, and terrain.
This priority-based projection produces temporally stable targets that are semantically compatible with the satellite view.
A lightweight BEV head predicts a semantic map $\operatorname{B}$ from the geospatial features $F_\mathrm{geo}$ and is supervised by $\hat{\operatorname{B}}$ using cross-entropy loss:
\begin{equation}
\mathcal{L}_{\mathrm{bev}} = \mathrm{CE}(\operatorname{B},\, \hat{\operatorname{B}}).
\end{equation}
This auxiliary objective encourages spatial consistency between 3D voxel reasoning and geospatial semantics.

OSM priors are used only to model \emph{static scene structure}, since time-invariant map information cannot reliably describe dynamic objects.
The suppression of dynamic categories is restricted to the geospatial branch and does not affect the main street-view prediction stream; dynamic classes therefore remain inferred from onboard observations.

\subsection{Training Loss}
In the GeoScene framework, we adopt the scene-class affinity loss from MonoScene~\cite{cao2022monoscene} to optimize precision, recall, and specificity concurrently.
The semantic scene completion loss is shown as follows:
\begin{equation}
    \mathcal{L}_{ssc} = \lambda_{s}\mathcal{L}^{sem}_{scal} + \lambda_{g}\mathcal{L}^{geo}_{scal} + \lambda_{ce}\mathcal{L}_{ce} +  \lambda_{d}\mathcal{L}_{d}.
\end{equation}
Following preliminary experiments, we set $\lambda_{s}$ = 1, $\lambda_{g}$ = 1, $\lambda_{ce}$ = 3 and $\lambda_{d}$ = 1, respectively.
The overall training loss function is formulated as follows: 
\begin{equation}
        \mathcal{L} = \mathcal{L}_{ssc} + \lambda_{bev}\mathcal{L}_{{bev}},
\end{equation}
where $\lambda_{bev}$ is set to 1.5. 

%% file: sec/4_exp.tex
\input{sec/tab/1semantickitti-val}

\input{sec/tab/2kitti360-test}

\section{Experiments}
\label{sec:exp}
We evaluate GeoScene on two outdoor SSC benchmarks: SemanticKITTI~\cite{behley2019semantickitti,Geiger2012kitti} and SSCBench-KITTI-360~\cite{li2023sscbench,Liao2022kitti360}. For reproducibility, the supplementary material provides complete documentation of the benchmark data and geospatial priors, including their sources, preprocessing, coordinate alignment, and class mappings, together with implementation settings, hyperparameters, and additional ablation, robustness, and qualitative experiments.
\subsection{Datasets}
\noindent\textbf{SemanticKITTI} contains 20 semantic classes and provides 10 training, 1 validation, and 11 test sequences. Because GPS information is unavailable for the official test split, we train on the GPS-available sequences, excluding sequence 03, and report results on the validation set.

\noindent\textbf{SSCBench-KITTI-360} contains 19 semantic classes with 7 training, 1 validation, and 1 test sequence. We follow its official split and report results on the test set. Both benchmarks use the same geospatial acquisition and alignment pipeline described below.

\input{sec/tab/3fair_comp}
\input{sec/tab/4ablation}
\input{sec/tab/ablation}
\noindent\textbf{Geospatial data and alignment.}
We retrieved satellite imagery from the Mapbox service~\cite{mapboxgljs} and queried OSM vector primitives through the Overpass interface~\cite{Olbricht2015,OpenStreetMap} on 27 January 2026. Mapbox imagery was accessed using a standard access token obtained by registering a free Mapbox account; no project-specific or privileged data service was required. For each frame, the benchmark GPS/IMU pose defines an ego-centered $51.2\,\mathrm{m}\times51.2\,\mathrm{m}$ region. The satellite crop and rasterized OSM map are resized to $512\times512$ pixels, yielding a processed ground-plane resolution of $0.1\,\mathrm{m}$ per pixel rather than the provider's native ground sampling distance. GPS coordinates are converted into a common metric map frame, translated relative to the ego position, and rotated according to the recorded heading to produce vehicle-aligned inputs. To support reproducibility, the supplementary material provides representative satellite crops and their rasterized OSM counterparts, the coordinate-transformation script used for alignment, and detailed retrieval and service-version metadata.

\subsection{Qualitative Results}

Figure~\ref{fig:figure4} qualitatively compares GeoScene with CGFormer and SGFormer on representative scenes from the SemanticKITTI validation set. Each row shows the driving image, satellite crop, rasterized OSM prior, predictions from the competing methods, our result, and the ground truth. The yellow circles highlight regions in which CGFormer produces missing, fragmented, or spatially inconsistent structures, while the red circles indicate corresponding failure regions in SGFormer; circles of the same color in the GeoScene results mark the locations used for direct comparison. Across these examples, GeoScene recovers more continuous road and sidewalk layouts, better preserves the boundaries of large static regions, and reduces isolated or fragmented predictions. The differences are particularly evident near road boundaries, intersections, and regions that are occluded or only weakly observed from the onboard view. These examples suggest that satellite appearance and structured OSM information provide complementary context for completing large-scale static structures beyond the camera field of view. The visual improvements are consistent with the class-wise results in Table~\ref{tab:1}, particularly for road, sidewalk, building, vegetation, and terrain.

\subsection{Quantitative Results}

Table~\ref{tab:1} compares GeoScene with existing methods on the SemanticKITTI validation set. GeoScene achieves the best overall performance, with 46.58\% IoU, 18.76\% mIoU, and 32.56\% Geo-mIoU. Compared with the strongest previously reported result for each metric, GeoScene improves IoU, mIoU, and Geo-mIoU by 0.18, 0.93, and 2.00 percentage points, respectively. The improvement is particularly evident for large-scale static classes, where GeoScene obtains the highest IoU for road, sidewalk, building, vegetation, and terrain. Notably, its building IoU reaches 30.1\%, surpassing the previous best result by 4.0 points. These results demonstrate the effectiveness of geospatial priors in improving both overall scene completion and the reconstruction of spatially persistent structures.

Table~\ref{tab:2} reports the results on the SSCBench-KITTI-360 test set. GeoScene again achieves the best aggregate performance, reaching 49.46\% IoU, 21.59\% mIoU, and 34.89\% Geo-mIoU. These scores exceed the strongest prior results by 1.27, 1.31, and 1.43 percentage points, respectively. GeoScene also consistently improves the reconstruction of large static regions, including road, building, fence, vegetation, and terrain. In particular, it increases building IoU from the previous best of 43.7\% to 47.9\%. The consistent improvements across both benchmarks confirm that the advantages of GeoScene generalize beyond a single dataset.

Table~\ref{tab:3} presents a controlled comparison with geospatially augmented baselines. For each method marked with $^\star$, we incorporate the same satellite imagery and OSM priors through a lightweight feature-fusion layer, while retaining the original baseline architecture. All variants are retrained using the same data split, geospatial preprocessing, and training protocol. Under this setting, GeoScene achieves the best performance, with 18.76\% mIoU and 32.56\% Geo-mIoU, demonstrating its stronger ability to exploit geospatial priors.

\subsection{Ablation Study}

\noindent\textbf{Ablation Study of Architecture Components.}
As shown in Table~\ref{tab:4}, introducing satellite imagery improves mIoU from 15.98\% to 16.32\%, demonstrating the benefit of complementary overhead appearance cues. Adding the structured OSM prior further increases mIoU to 16.67\%, although the IoU remains nearly unchanged, suggesting that OSM primarily strengthens semantic identification before adaptive fusion is applied. With both geospatial inputs, independently enabling the observation and geospatial weighting branches yields 17.21\% and 17.48\% mIoU, respectively. The larger semantic gain from the geospatial branch indicates the value of structured priors in weakly observed regions, whereas the observation branch helps preserve evidence from the onboard view. Combining both weights improves the results to 17.87\% mIoU and 45.32\% IoU, exceeding either individual branch and confirming their complementary roles. Finally, WGVR provides additional gains of 0.89 mIoU and 1.26 IoU points, reaching 18.76\% mIoU and 46.58\% IoU. Overall, the complete model improves the baseline by 2.78 mIoU and 2.45 IoU points, showing that geospatial inputs, dual-prior weighting, and voxel refinement contribute progressively, with WGVR producing the largest gain in geometric completion.

\paragraph{Ablation and robustness analysis.}
As shown in Tables~\ref{tab:osm}--\ref{tab:he}, combining road and building priors yields the best performance, reaching 32.56 Geo-mIoU, 18.76 mIoU, and 46.58 IoU. The road-only configuration consistently outperforms the building-only configuration, while their combination provides further gains across all three metrics, confirming that road topology and building geometry offer complementary structural cues. The perturbation experiments further show that GeoScene degrades progressively rather than catastrophically when GPS/IMU accuracy deteriorates. Under mild errors of $0.5\,\mathrm{m}$ or $1^\circ$, mIoU decreases by only 0.53 and 0.54 points, respectively. Even with severe perturbations of $5\,\mathrm{m}$ or $10^\circ$, the model retains 16.17 and 15.12 mIoU, together with 42.15 and 41.30 IoU. These results indicate that inaccurate geospatial alignment gradually reduces the benefit of external priors, while the onboard observation pathway provides a fallback that helps avoid abrupt failure under degraded localization.

\section{Conclusion}
\label{sec:conclusion}
We presented GeoScene, a geospatial-prior-guided framework for 3D semantic scene completion beyond the onboard field of view.
Rather than treating external information as a hard constraint, GeoScene combines satellite imagery and structured OpenStreetMap cues through learned voxel-wise soft reliability weights.
The Dual-Priors Weighted Classifier estimates the complementary reliability of onboard observations and geospatial guidance, while the Weights-Guided Voxel Refiner uses these weights to adaptively integrate the two feature sources.
Experiments on SemanticKITTI and SSCBench-KITTI-360 demonstrate state-of-the-art performance under the geospatial-prior-assisted setting.


%% file: sec/tab/1semantickitti-val.tex

\begin{table*}[ht]
  \centering
  \small
  \setlength{\tabcolsep}{2pt}

  \begin{tabular}{l|ccc|rrrrrrrrrrrrrrrrrrr}
    \toprule
    \textbf{Methods}&\rotatebox{90}{\textbf{IoU}}& \rotatebox{90}{\textbf{mIoU}}   & \rotatebox{90}{\textbf{Geo-mIoU}}
    & \rotatebox{90}{\vcenteredbox{\colorbox[RGB]{255,0,255}{\textcolor[RGB]{255,0,255}{\rule{1px}{1px}}}} \textbf{road{$^\dagger$}}} 
    & \rotatebox{90}{\vcenteredbox{\colorbox[RGB]{75,0,75}{\textcolor[RGB]{75,0,75}{\rule{1px}{1px}}}} \textbf{sidewalk{$^\dagger$}}} 
    & \rotatebox{90}{\vcenteredbox{\colorbox[RGB]{255,150,255}{\textcolor[RGB]{255,150,255}{\rule{1px}{1px}}}} \textbf{parking{$^\dagger$}}} 
    & \rotatebox{90}{\vcenteredbox{\colorbox[RGB]{175,0,75}{\textcolor[RGB]{175,0,75}{\rule{1px}{1px}}}} \textbf{other-grnd{$^\dagger$}}} 
    &  \rotatebox{90}{\vcenteredbox{\colorbox[RGB]{255,200,0}{\textcolor[RGB]{255,200,0}{\rule{1px}{1px}}}} \textbf{building{$^\dagger$}}} 
    &  \rotatebox{90}{\vcenteredbox{\colorbox[RGB]{100,150,245}{\textcolor[RGB]{100,150,245}{\rule{1px}{1px}}}} \textbf{car}} 
    & \rotatebox{90}{\vcenteredbox{\colorbox[RGB]{80,30,180}{\textcolor[RGB]{80,30,180}{\rule{1px}{1px}}}} \textbf{truck}} 
    & \rotatebox{90}{\vcenteredbox{\colorbox[RGB]{100,230,245}{\textcolor[RGB]{100,230,245}{\rule{1px}{1px}}}} \textbf{bicycle}}  
    & \rotatebox{90}{\vcenteredbox{\colorbox[RGB]{30,60,150}{\textcolor[RGB]{30,60,150}{\rule{1px}{1px}}}} \textbf{motorcycle}} 
    & \rotatebox{90}{\vcenteredbox{\colorbox[RGB]{0,0,255}{\textcolor[RGB]{0,0,255}{\rule{1px}{1px}}}} \textbf{other-vehicle}} 
    & \rotatebox{90}{\vcenteredbox{\colorbox[RGB]{0,175,0}{\textcolor[RGB]{0,175,0}{\rule{1px}{1px}}}} \textbf{vegetation{$^\dagger$}}} 
    & \rotatebox{90}{\vcenteredbox{\colorbox[RGB]{135,60,0}{\textcolor[RGB]{135,60,0}{\rule{1px}{1px}}}}  \textbf{trunk}} 
    & \rotatebox{90}{\vcenteredbox{\colorbox[RGB]{150,240,80}{\textcolor[RGB]{150,240,80}{\rule{1px}{1px}}}} \textbf{terrain{$^\dagger$}}} 
    & \rotatebox{90}{\vcenteredbox{\colorbox[RGB]{255,30,30}{\textcolor[RGB]{255,30,30}{\rule{1px}{1px}}}} \textbf{person}} 
    & \rotatebox{90}{\vcenteredbox{\colorbox[RGB]{255,40,200}{\textcolor[RGB]{255,40,200}{\rule{1px}{1px}}}} \textbf{bicyclist}} 
    & \rotatebox{90}{\vcenteredbox{\colorbox[RGB]{150,30,90}{\textcolor[RGB]{150,30,90}{\rule{1px}{1px}}}}  \textbf{motorcyclist}} 
    &  \rotatebox{90}{\vcenteredbox{\colorbox[RGB]{255,120,50}{\textcolor[RGB]{255,120,50}{\rule{1px}{1px}}}} \textbf{fence}} 
    & \rotatebox{90}{\vcenteredbox{\colorbox[RGB]{255,240,150}{\textcolor[RGB]{255,240,150}{\rule{1px}{1px}}}} \textbf{pole}} 
    & \rotatebox{90}{\vcenteredbox{\colorbox[RGB]{255,0,0}{\textcolor[RGB]{255,0,0}{\rule{1px}{1px}}}} \textbf{traf.-sign}} \\
    
    \midrule
MonoScene\textsubscript{\textcolor{cyan}{\scalebox{0.7}{[CVPR'22]}}} &
36.86&11.08&22.80&
56.5&26.7&14.3&0.5&14.1&23.3&7.0&0.6&0.5&1.5&17.9&2.8&29.6&1.9&1.2&0.0&5.8&4.1&2.3 \\

VoxFormer\textsubscript{\textcolor{cyan}{\scalebox{0.7}{[CVPR'23]}}}&
44.15&13.35&25.56&
53.6&26.5&19.7&0.4&19.5&26.5&7.3&1.3&0.6&7.8&26.1&6.1&33.1&1.9&2.0&0.0&7.3&9.2&4.9\\

Symphonize\textsubscript{\textcolor{cyan}{\scalebox{0.7}{[CVPR'24]}}}&
41.92&14.89&25.49&
56.4&27.6&15.3&1.0&21.6&28.7&20.4&2.5&2.8&13.9&25.7&6.6&30.9&3.5&2.2&0.0&8.4&9.6&5.8\\

H2GFormer\textsubscript{\textcolor{cyan}{\scalebox{0.7}{[AAAI'24]}}}&
44.69&14.29&27.52&
57.0&29.4&21.7&0.3&20.5&28.2&6.8&1.0&0.9&9.3&27.4&7.8&36.3&1.2&0.1&0.0&8.0&9.9&5.8\\

CGFormer\textsubscript{\textcolor{cyan}{\scalebox{0.7}{[NIPS'24]}}}&
45.99&16.87&29.83&
65.5&32.3&20.8&0.2&23.5&
\textbf{34.3}&19.4&
\textbf{4.6}&2.7&7.7&
26.9&8.8&39.5&
2.4&\textbf{4.1}&0.0&
9.2&10.7&\underline{7.8}\\

VLScene\textsubscript{\textcolor{cyan}{\scalebox{0.7}{[AAAI'25]}}}&
44.69&\underline{17.83}&\underline{29.29}&
63.1&31.1&
\textbf{24.4}&
0.2&24.9&
33.4&
\underline{30.7}&
1.8&
\textbf{3.6}&
\textbf{18.3}&
26.0&
8.1&
35.3&
\textbf{4.3}&
\underline{2.6}&0.0&
\textbf{12.1}&
\textbf{11.9}&
6.3\\

SGFormer\textsubscript{\textcolor{cyan}{\scalebox{0.7}{[CVPR'25]}}}&
45.01&16.68&29.35&
60.9&31.5&19.3&
\textbf{2.1}&
\underline{26.1}&
32.2&20.3&3.0&
\underline{3.1}&
11.6&
27.1&8.3&38.5&
\underline{3.7}&
1.5&0.0&
9.3&{11.6}&7.2\\

Ocean\textsubscript{\textcolor{cyan}{\scalebox{0.7}{[AAAI'26]}}}&
\underline{46.40}&17.39&\underline{30.56}&
\underline{66.1}&
\underline{34.3}&
\underline{21.9}&
0.1&23.4&
\underline{34.0}&
19.3&
\underline{3.2}&
2.0&
\underline{15.3}&
\underline{28.1}&
\underline{8.9}&
\underline{40.0}&
\underline{3.7}&
1.2&0.0&
10.3&10.8&
\underline{7.8}\\

\hline

\rowcolor{gray!20}
\textbf{GeoScene}&
\textbf{46.58}&
\textbf{18.76}&
\textbf{32.56}&
\textbf{66.7}&
\textbf{36.2}&
\underline{22.3}&
\underline{1.8}&
\textbf{30.1}&
\underline{34.0}&
\textbf{32.3}&
1.1&
2.9&
13.9&
\textbf{30.3}&
\textbf{9.1}&
\textbf{40.5}&
2.7&
2.1&
0.0&
\underline{10.8}&
\underline{11.8}&
\textbf{7.9}\\
    \bottomrule
  \end{tabular}
  \caption{Quantitative results on the SemanticKITTI validation set. The best and second-best results are shown in \textbf{bold} and \underline{underlined}, respectively. $\dagger$ represents the geospatial category selection used by SGFormer.}
  \label{tab:1}

\end{table*}


%% file: sec/tab/2kitti360-test.tex
\begin{table*}[ht]
  \centering
  \small
  \setlength{\tabcolsep}{2pt}
  
  \begin{tabular}{l|ccc|rrrrrrrrrrrrrrrrrr}
    \toprule
    \textbf{Methods}&\rotatebox{90}{\textbf{IoU}} &\rotatebox{90}{\textbf{mIoU}}&\rotatebox{90}{\textbf{Geo-mIoU}}
    &  \rotatebox{90}{\vcenteredbox{\colorbox[RGB]{100,150,245}{\textcolor[RGB]{100,150,245}{\rule{1px}{1px}}}} \textbf{car}} 
    & \rotatebox{90}{\vcenteredbox{\colorbox[RGB]{100,230,245}{\textcolor[RGB]{100,230,245}{\rule{1px}{1px}}}} \textbf{bicycle}}  
    & \rotatebox{90}{\vcenteredbox{\colorbox[RGB]{30,60,150}{\textcolor[RGB]{30,60,150}{\rule{1px}{1px}}}} \textbf{motocycle}} 
    & \rotatebox{90}{\vcenteredbox{\colorbox[RGB]{80,30,180}{\textcolor[RGB]{80,30,180}{\rule{1px}{1px}}}} \textbf{truck}} 
    & \rotatebox{90}{\vcenteredbox{\colorbox[RGB]{0,0,255}{\textcolor[RGB]{0,0,255}{\rule{1px}{1px}}}} \textbf{other-vehicle}} 
    & \rotatebox{90}{\vcenteredbox{\colorbox[RGB]{255,30,30}{\textcolor[RGB]{255,30,30}{\rule{1px}{1px}}}} \textbf{person}} 
    &  \rotatebox{90}{\vcenteredbox{\colorbox[RGB]{255,0,255}{\textcolor[RGB]{255,0,255}{\rule{1px}{1px}}}} \textbf{road{$^\dagger$}}} 
    & \rotatebox{90}{\vcenteredbox{\colorbox[RGB]{255,150,255}{\textcolor[RGB]{255,150,255}{\rule{1px}{1px}}}} \textbf{parking{$^\dagger$}}} 
    & \rotatebox{90}{\vcenteredbox{\colorbox[RGB]{75,0,75}{\textcolor[RGB]{75,0,75}{\rule{1px}{1px}}}} \textbf{sidewalk{$^\dagger$}}} 
    & \rotatebox{90}{\vcenteredbox{\colorbox[RGB]{175,0,75}{\textcolor[RGB]{175,0,75}{\rule{1px}{1px}}}} \textbf{other-grnd{$^\dagger$}}} 
    &  \rotatebox{90}{\vcenteredbox{\colorbox[RGB]{255,200,0}{\textcolor[RGB]{255,200,0}{\rule{1px}{1px}}}} \textbf{building{$^\dagger$}}} 
    &  \rotatebox{90}{\vcenteredbox{\colorbox[RGB]{255,120,50}{\textcolor[RGB]{255,120,50}{\rule{1px}{1px}}}} \textbf{fence}} 
    & \rotatebox{90}{\vcenteredbox{\colorbox[RGB]{0,175,0}{\textcolor[RGB]{0,175,0}{\rule{1px}{1px}}}} \textbf{vegetation{$^\dagger$}}} 
    & \rotatebox{90}{\vcenteredbox{\colorbox[RGB]{150,240,80}{\textcolor[RGB]{150,240,80}{\rule{1px}{1px}}}} \textbf{terrain{$^\dagger$}}} 
    & \rotatebox{90}{\vcenteredbox{\colorbox[RGB]{255,240,150}{\textcolor[RGB]{255,240,150}{\rule{1px}{1px}}}} \textbf{pole}} 
    & \rotatebox{90}{\vcenteredbox{\colorbox[RGB]{255,0,0}{\textcolor[RGB]{255,0,0}{\rule{1px}{1px}}}} \textbf{traf.-sign}} 
    & \rotatebox{90}{\vcenteredbox{\colorbox[RGB]{0, 150, 255}{\textcolor[RGB]{0, 150, 255}{\rule{1px}{1px}}}}  \textbf{other-struct.{$^\dagger$}}} 
    & \rotatebox{90}{\vcenteredbox{\colorbox[RGB]{255, 255, 50}{\textcolor[RGB]{255, 255, 50}{\rule{1px}{1px}}}} \textbf{other-obj.}}   \\
    
    \midrule
MonoScene\textsubscript{\textcolor{cyan}{\scalebox{0.7}{[CVPR'22]}}}&37.87&12.31&21.40&19.3&0.4&0.6&8.0&2.0&0.9&48.4&11.4&28.1&3.3&32.9&3.5&26.2&16.8&6.9&5.7&4.2&3.1\\
VoxFormer\textsubscript{\textcolor{cyan}{\scalebox{0.7}{[CVPR'23]}}}&38.76&11.91&20.68&17.8&1.2&0.9&4.6&2.1&1.6&47.0&9.7&27.2&2.9&31.2&5.0&29.0&14.7&6.5&6.9&3.8&2.4\\
OccFormer\textsubscript{\textcolor{cyan}{\scalebox{0.7}{[ICCV'23]}}}&40.27&13.81&24.59&22.6&0.7&0.3&9.9&3.8&2.8&54.3&13.4&31.5&3.6&36.4&4.8&31.0&19.5&7.8&8.5&7.0&4.6\\
Symphonies\textsubscript{\textcolor{cyan}{\scalebox{0.7}{[CVPR'24]}}}&44.12&18.58&25.98&\textbf{30.0}&1.9&\underline{5.9}&\textbf{25.1}&\textbf{12.1}&\underline{8.2}&54.9&13.8&32.8&\underline{6.9}&35.1&8.6&38.3&11.5&14.0&9.6&\textbf{14.4}&\textbf{11.3}\\
CGFormer\textsubscript{\textcolor{cyan}{\scalebox{0.7}{[NIPS'24]}}}&48.07&20.05&30.49&\underline{29.9}&3.4&4.0&17.6&6.8&6.6&\underline{63.9}&17.2&\underline{40.7}&5.5&42.7&8.2&38.8&\underline{24.9}&16.2&17.5&10.2&6.8\\
VLScene\textsubscript{\textcolor{cyan}{\scalebox{0.7}{[AAAI'25]}}}&46.08&19.10&32.27&29.0&\underline{4.7}&\textbf{7.7}&18.3&7.6&7.4&60.1&\underline{17.4}&39.0&6.0&42.1&\underline{9.6}&36.5&24.8&\textbf{17.0}&\underline{18.8}&10.5&6.5\\
SGFormer\textsubscript{\textcolor{cyan}{\scalebox{0.7}{[CVPR'25]}}}&46.35&18.30&29.01&27.8&0.9&2.6&10.7&5.7&4.3&61.0&13.2&37.0&5.1&43.1&7.5&39.0&\underline{24.9}&15.8&16.9&8.9&5.3\\
EGS\textsubscript{\textcolor{cyan}{\scalebox{0.7}{[TIP'26]}}}& \underline{48.61}& 19.74& 29.89 & \textbf{30.4}&2.0&3.6&12.6&\underline{9.5}&\textbf{14.3}&63.4&\textbf{34.3}&37.6&5.1&34.4&\underline{9.8}&32.1&20.4&\textbf{17.0}&9.7&11.8&7.3 \\
Ocean\textsubscript{\textcolor{cyan}{\scalebox{0.7}{[AAAI'26]}}}&\underline{48.19}&\underline{20.28}&\underline{33.46}&29.3&3.7&4.6&15.1&7.7&6.8&63.7&17.0&\textbf{40.9}&5.0&\underline{43.7}&8.9&\underline{39.2}&24.7&\underline{16.7}&\textbf{19.2}&10.8&\underline{8.3}\\
\hline
\rowcolor{gray!20}\textbf{GeoScene}& \textbf{49.46} & \textbf{21.59} & \textbf{34.89} & 29.7 & \textbf{5.1} & 3.6 & \underline{22.4} & \underline{8.8} & \textbf{8.3} & \textbf{64.9} & \textbf{17.5} & 39.9 & \textbf{7.2} & \textbf{47.9} & \textbf{10.0} & \textbf{41.5} & \textbf{25.3} & \underline{16.7} & 18.4 & \underline{13.5} & 7.9\\
    \bottomrule
  \end{tabular}
  \caption{Quantitative results on the SSCBench-KITTI-360 test set. The best and second-best results are shown in \textbf{bold} and \underline{underlined}, respectively. $\dagger$ represents the geospatial category selection used by SGFormer.}
  \label{tab:2}

\end{table*}

%% file: sec/tab/3fair_comp.tex
\begin{table}[t]
  \centering
  \small
  \setlength{\tabcolsep}{3pt}
  \begin{tabular}{@{}lcccc@{}}
    \toprule
    \textbf{Method} & \textbf{mIoU}& \textbf{Geo-mIoU} & \textbf{Time (s)} & \textbf{Params (M)} \\
    \midrule
    MonoScene$^\star$  & 12.22 & 23.12 & 0.291 & 136.2 \\
    VoxFormer$^\star$ & 13.88 & 25.84 & 0.278 & 61.7 \\
    BRGScene$^\star$ & 16.37 &28.53 & 0.306 & 165.2 \\
    CGFormer$^\star$ & 17.12& 29.61 & 0.225 & 125.8 \\
    VLScene$^\star$ & 17.63 & 30.13 & 0.235 & 51.2 \\
    SGFormer & 16.68 & 29.35& -- & 126.9 \\
    \rowcolor{gray!20}\textbf{GeoScene} & \textbf{18.76} & \textbf{32.56} & \textbf{0.186} & 82.0 \\
    \bottomrule
  \end{tabular}
  \caption{Comparison of accuracy and computational efficiency. $^\star$ denotes existing methods augmented with the same geospatial priors for a fair comparison.}
  \label{tab:3}
\end{table}

%% file: sec/tab/4ablation.tex
\begin{table}[t]
  \centering
  \small
  \setlength{\tabcolsep}{4pt}
  \begin{tabular}{@{}c|cc|cc|c|cc@{}}
    \toprule
    \multirow{2}{*}{\textbf{Variant}} & \multirow{2}{*}{\textbf{Sat.}} & \multirow{2}{*}{\textbf{OSM}}
    & \multicolumn{2}{c|}{\textbf{DPWC}} & \multirow{2}{*}{\textbf{WGVR}}
    & \multirow{2}{*}{\textbf{IoU}} & \multirow{2}{*}{\textbf{mIoU}} \\
    & & & $W_\mathrm{Obs}$ & $W_\mathrm{Geo}$ & & & \\
    \midrule
    Base & & & & & & 44.13 & 15.98 \\
    1 & \checkmark & & & & & 44.35 & 16.32 \\
    2 & \checkmark & \checkmark & & & & 44.26 & 16.67 \\
    3 & \checkmark & \checkmark & \checkmark & & & 44.58 & 17.21 \\
    4 & \checkmark & \checkmark & & \checkmark & & 44.13 & 17.48 \\
    5 & \checkmark & \checkmark & \checkmark & \checkmark & & 45.32 & 17.87 \\
    \rowcolor{gray!20}6 & \checkmark & \checkmark & \checkmark & \checkmark & \checkmark & \textbf{46.58} & \textbf{18.76} \\
    \bottomrule
  \end{tabular}
  \caption{Ablation study of the architecture components.}
  \label{tab:4}
\end{table}

%% file: sec/tab/ablation.tex
\begin{table*}[t]
  \centering
  \small

  \begin{minipage}[t]{0.38\textwidth}
    \centering
    \setlength{\tabcolsep}{2pt}
    \begin{tabular}{@{}ccccc@{}}
      \toprule
      \textbf{Building} & \textbf{Road} & \textbf{Geo-mIoU} & \textbf{mIoU} & \textbf{IoU} \\
      \midrule
      \ding{51} & \ding{55} & 30.93 & 17.13 & 45.05 \\
      \ding{55} & \ding{51} & 31.46 & 17.97 & 45.47 \\
      \ding{51} & \ding{51} & \textbf{32.56} & \textbf{18.76} & \textbf{46.58} \\
      \bottomrule
    \end{tabular}
    \caption{Ablation of OSM categories.}
    \label{tab:osm}
  \end{minipage}
  \hfill
  \begin{minipage}[t]{0.29\textwidth}
    \centering
    \setlength{\tabcolsep}{1.2pt}
    \begin{tabular}{@{}cccc@{}}
      \toprule
      \textbf{Shift} & \textbf{Geo-mIoU} & \textbf{mIoU} & \textbf{IoU} \\
      \midrule
      0 m & \textbf{32.56} & \textbf{18.76} & \textbf{46.58} \\
      0.5 m & 31.31 & 18.23 & 45.99 \\
      1 m & 30.71 & 17.99 & 45.64 \\
      2 m & 28.96 & 17.31 & 44.74 \\
      5 m & 26.32 & 16.17 & 42.15 \\
      \bottomrule
    \end{tabular}
    \caption{Satellite translation error.}
    \label{tab:te}
  \end{minipage}
  \hfill
  \begin{minipage}[t]{0.29\textwidth}
    \centering
    \setlength{\tabcolsep}{1.2pt}
    \begin{tabular}{@{}cccc@{}}
      \toprule
      \textbf{Angle} & \textbf{Geo-mIoU} & \textbf{mIoU} & \textbf{IoU} \\
      \midrule
      $0^\circ$ & \textbf{32.56} & \textbf{18.76} & \textbf{46.58} \\
      $1^\circ$ & 31.24 & 18.22 & 45.98 \\
      $2^\circ$ & 30.35 & 17.85 & 45.57 \\
      $5^\circ$ & 27.08 & 16.54 & 43.90 \\
      $10^\circ$ & 25.14 & 15.12 & 41.30 \\
      \bottomrule
    \end{tabular}
    \caption{Satellite heading error.}
    \label{tab:he}
  \end{minipage}
\end{table*}

%% file: sec/X_suppl.tex
\section{Additional Implementation and Experimental Details}
\label{app:details}

This appendix provides implementation details and supporting analyses omitted from the main paper because of space constraints. It covers geospatial preprocessing, module-level ablations, robustness tests, additional qualitative results, and representative failure cases. Benchmark descriptions and aggregate results reported in the main text are not repeated.

\subsection{Geospatial Data Preparation}

\paragraph{Sources and temporal metadata.}
We retrieved the satellite crops from the Mapbox satellite imagery service and queried OpenStreetMap (OSM) vectors through the Overpass interface at \texttt{2026-01-27T00:00:00Z} (27 January 2026, UTC). Access to Mapbox used a standard token obtained by registering a free Mapbox account; the token itself is a private credential and is therefore not included in the supplementary material. The satellite crops correspond to the imagery served by Mapbox at retrieval time, because the interface does not expose acquisition timestamps for individual source images or a fixed native-imagery version. Likewise, the OSM data represent the database state returned by Overpass at the recorded retrieval time rather than a separately named snapshot. We provide the retrieval timestamp and applicable service metadata so that the provenance and temporal scope of the geospatial inputs are explicit.

\paragraph{Crop scale and processed resolution.}
For each frame, the benchmark GPS/IMU pose defines an ego-centred region of $51.2\,\mathrm{m}\times51.2\,\mathrm{m}$. We resize both the satellite crop and the rasterized OSM map to $512\times512$ pixels. The resulting ground-plane resolution is $0.1\,\mathrm{m}$ per pixel. This value describes the processed input rather than the native ground sampling distance of the Mapbox imagery. Both benchmarks use the same crop extent and output resolution.

\paragraph{GPS/IMU-to-map alignment.}
Let $\mathbf{p}_{\mathrm{map}}$ denote a satellite pixel or OSM primitive in a common metric map coordinate system. The vehicle position and heading from the benchmark GPS/IMU record are denoted by $\mathbf{p}_{\mathrm{ego}}$ and $\theta_{\mathrm{ego}}$, respectively. We transform map coordinates into the vehicle-aligned ground plane as
\begin{equation}
\mathbf{p}_{\mathrm{veh}}=
\mathbf{R}(-\theta_{\mathrm{ego}})
\left(\mathbf{p}_{\mathrm{map}}-\mathbf{p}_{\mathrm{ego}}\right),
\end{equation}
where $\mathbf{R}(\cdot)$ denotes a two-dimensional rotation matrix. We then crop both geospatial sources around the vehicle and rasterize them on the SSC grid.

\paragraph{OSM mapping and missing coverage.}
Following the procedural parsing strategy, we assign road-related primitives to \textit{road} and closed building footprints to \textit{building}. Unsupported or unmapped locations are assigned to \textit{others}. These coarse categories serve only as soft auxiliary priors. An all-zero map represents a crop with no valid road or building primitive, whereas partially covered crops retain the available primitives. We do not use ground-truth semantic labels to fill missing map regions. Predictions in these regions therefore rely primarily on onboard observations.

\paragraph{Geo-mIoU class definition.}
Geo-mIoU evaluates semantic completion over large-scale static classes that can be meaningfully associated with geospatial structure. Following the class selection used by SGFormer, the SemanticKITTI subset is \textbf{road},\textbf{sidewalk},\textbf{parking},\textbf{other-ground},\textbf{building},\textbf{vegetation},\textbf{terrain}, whereas the SSCBench-KITTI-360 subset is \textbf{road},\textbf{parking},\textbf{sidewalk},\textbf{other-ground},\textbf{building},\textbf{vegetation},\textbf{terrain},\textbf{other-structure}.

For each benchmark, Geo-mIoU is the unweighted mean of the class-wise IoU values over the corresponding subset. This evaluation subset is distinct from the three-category OSM input representation, which contains only \{\textit{road}, \textit{building}, \textit{others}\}.

\subsection{Training Configuration}

We optimize GeoScene with AdamW using an initial learning rate of $10^{-4}$ and a weight decay of $0.01$. A multi-step schedule controls the learning rate. Training proceeds for 24 epochs with a batch size of 4 on two NVIDIA A100 GPUs with 80 GB memory. The SSC volume spans $51.2\,\mathrm{m}\times51.2\,\mathrm{m}\times6.4\,\mathrm{m}$ and contains $256\times256\times32$ voxels. SemanticKITTI and SSCBench-KITTI-360 use 20 and 19 output classes, respectively. Unless stated otherwise, we set $r=1.5$ and $o=1.0$ in the class-aware geospatial scaling term of Eq.~(4). This setting assigns greater weight to road and building priors than to other regions. We set the fusion coefficient in Eq.~(12) to $\alpha=0.7$, thereby placing greater emphasis on the directly weighted fusion $V_{\mathrm{w}}$ than on the normalized fusion $V_{\mathrm{norm}}$. We additionally set $\lambda_{\mathrm{bev}}=1.5$ and evaluate its sensitivity below.

\subsection{Detailed Module Architecture}

\begin{figure*}[t]
  \centering
  \includegraphics[width=\textwidth]{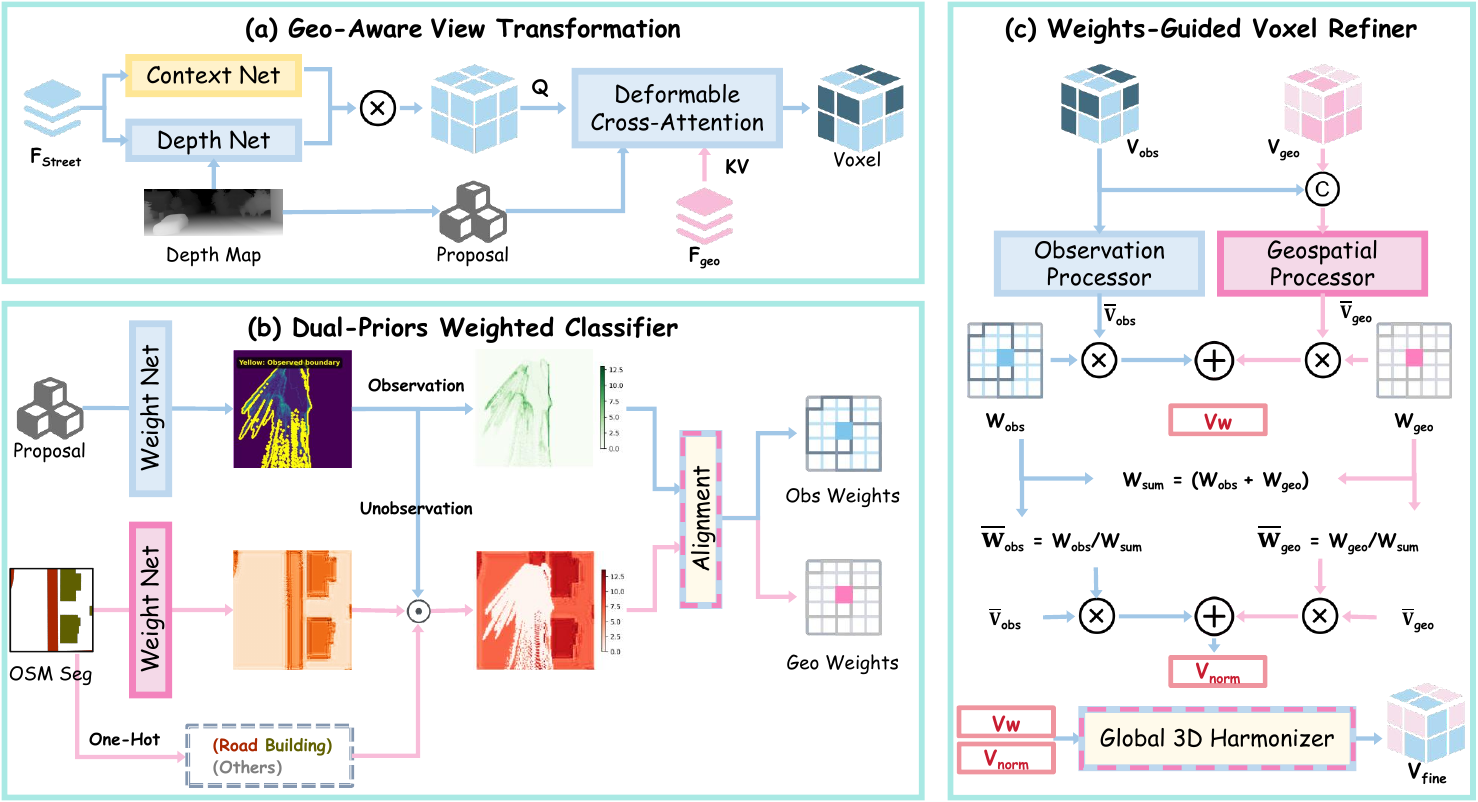}
  \caption{Detailed architecture of GeoScene. (a) Geo-Aware View Transformation projects observation and geospatial features into the voxel representation. (b) The Dual-Priors Weighted Classifier learns separate soft reliability fields for observation and geospatial cues. (c) The Weights-Guided Voxel Refiner uses these fields to refine and fuse the two feature streams.}
  \label{fig:detailed_modules}
\end{figure*}
Figure~\ref{fig:detailed_modules} complements the overview in Figure~\ref{fig:figure2} by detailing the information flow within each module. The observation stream preserves geometry supported by onboard evidence. The geospatial stream provides satellite and OSM cues for weakly observed regions. DPWC estimates the soft reliability fields that control the voxel-wise contributions of both streams in WGVR.

\section{Additional Ablation Studies}
\label{app:ablations}
\input{sec/tab/supp_tab}
\begin{table*}[t]
  \centering
  \footnotesize
  \begin{tabular}{@{}lcccccc@{}}
    \toprule
    \textbf{Drop ratio} & 0\% & 10\% & 20\% & 40\% & \textbf{SGFormer} & \textbf{CGFormer}\\
    \midrule
    \textbf{mIoU (\%)} & 18.76 & 18.37 & 18.04 & 17.51 & 16.68 & 16.87\\
    \bottomrule
  \end{tabular}
  \caption{Inference-time geospatial dropout without retraining, with SGFormer and CGFormer included as reference baselines.}
  \label{tab:geo_dropout}
\end{table*}
\begin{figure*}[t]
  \centering
  \includegraphics[width=\linewidth]{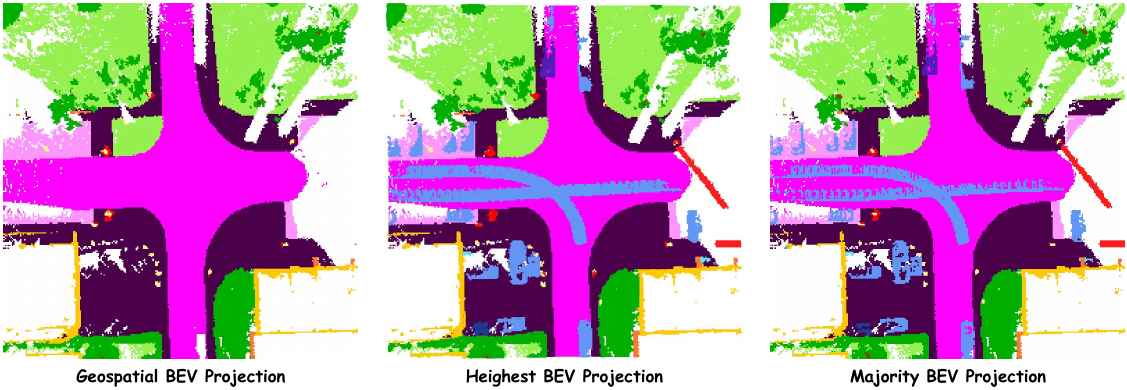}
  \caption{Qualitative comparison of BEV target projection strategies. The geospatial-priority strategy retains stable static structures while suppressing time-inconsistent labels.}
  \label{fig:bev_projection}
\end{figure*}

\paragraph{Complementarity of geospatial sources.}
Table~\ref{tab:view} isolates the contribution of each geospatial source. Adding satellite imagery to the camera-only model increases Geo-mIoU by 1.88 points. Adding OSM alone yields a larger gain of 4.98 points. Combining both sources achieves 32.56\% Geo-mIoU, 18.76\% mIoU, and 46.58\% IoU. These results show that satellite appearance and structured map semantics provide complementary cues.

\begin{figure}[t]
  \centering
  \includegraphics[width=\linewidth]{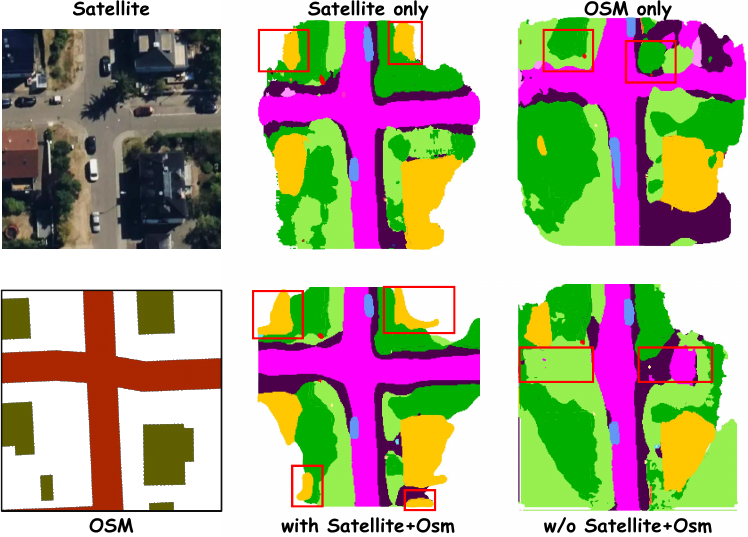}
  \caption{Qualitative ablation of geospatial inputs. Satellite imagery provides coarse appearance and layout cues, whereas OSM specifies road and building structure. Combining both priors improves completion in occluded and weakly observed regions.}
  \label{fig:geo_input_ablation}
\end{figure}

Figure~\ref{fig:geo_input_ablation} shows the corresponding qualitative differences. Satellite-only predictions recover the broad scene layout but retain ambiguous boundaries. OSM-only predictions better preserve roads and buildings but lack appearance context. Combining both sources improves the continuity of large static structures and their agreement with the reference layout.

\paragraph{BEV supervision strength.}
Table~\ref{tab:loss} examines the strength of BEV supervision. Performance peaks at $\lambda_{\mathrm{bev}}=1.5$ and decreases slightly at larger values. Excessive BEV supervision may therefore compete with voxel-level SSC supervision.

\paragraph{Class-aware scaling coefficients.}
Table~\ref{tab:ro_sensitivity} evaluates the coefficients $r$ and $o$, which control the relative contribution of road/building priors and other regions. Assigning a larger weight to road and building regions consistently improves performance over the reversed setting $(r,o)=(0.5,2.0)$. The default setting $(1.5,1.0)$ achieves the best results, reaching 32.56\% Geo-mIoU, 18.76\% mIoU, and 46.58\% IoU. Increasing $r$ further to 3.0 reduces the three metrics by 0.24, 0.53, and 0.35 points, respectively. These results suggest that emphasizing structured static categories is beneficial, whereas excessive scaling can suppress complementary information from other regions.

\paragraph{Fusion coefficient.}
Table~\ref{tab:alpha_sensitivity} examines the balance between the directly weighted fusion $V_{\mathrm{w}}$ and the normalized fusion $V_{\mathrm{norm}}$. Performance improves as $\alpha$ increases from 0.3 to 0.7, with Geo-mIoU, mIoU, and IoU increasing by 1.23, 0.77, and 0.55 points, respectively. The default value $\alpha=0.7$ achieves the best performance across all three metrics. Increasing $\alpha$ to 0.9 causes a moderate decline, indicating that retaining a contribution from the normalized fusion is important for balancing the two feature streams.

\paragraph{Soft reliability modeling.}
Table~\ref{tab:wei} compares hard gating, generic soft weighting, and the proposed Dual-Priors Weighted Classifier (DPWC). Hard gating restricts adaptive feature modulation. Generic soft weighting does not distinguish observation reliability from geospatial guidance. DPWC performs best by learning a separate soft reliability field for each source.

\paragraph{Voxel refinement.}
Table~\ref{tab:wgvr} compares WGVR with concatenation, addition, and attention-based fusion. WGVR achieves the highest scores across the three metrics. This result supports weight-guided dual-branch refinement over a single generic fusion operator.

\paragraph{BEV target construction.}
Table~\ref{tab:pro} compares highest-point, majority, and geospatial-priority projections. The proposed strategy prioritizes stable static structures and suppresses time-inconsistent dynamic labels. It consequently produces BEV targets that are more compatible with overhead geospatial inputs. Figure~\ref{fig:bev_projection} illustrates the differences among the three strategies.

\section{Robustness and Class-wise Analysis}
\label{app:robustness}

\paragraph{Spatial misalignment.}
Tables~\ref{tab:te} and~\ref{tab:he} assess sensitivity to translation and heading errors. Performance declines progressively as either perturbation increases. Errors of $0.5\,\mathrm{m}$ or $1^\circ$ have a limited effect, whereas larger perturbations reduce Geo-mIoU more markedly. GeoScene therefore tolerates mild localization noise but remains sensitive to severe map--vehicle misalignment.

\paragraph{Inference-time geospatial dropout.}
We further assess missing geospatial inputs without retraining. At inference time, satellite and OSM inputs are randomly replaced with zeros. As shown in Table~\ref{tab:geo_dropout}, mIoU decreases from 18.76 at 0\% dropout to 18.37, 18.04, and 17.51 at dropout ratios of 10\%, 20\%, and 40\%, respectively. This gradual decline suggests that the observation branch provides a fallback when external priors are intermittently unavailable. The experiment uses a separate robustness checkpoint and is reported independently of the final full-model results.

\paragraph{Dynamic classes.}
Static geospatial priors are confined to the auxiliary geospatial branch and do not replace observation-driven predictions for dynamic objects. On SemanticKITTI, GeoScene increases the IoU of car, truck, other-vehicle, and bicyclist from 32.2, 20.3, 11.6, and 1.5 with SGFormer to 34.0, 32.3, 13.9, and 2.1, respectively. These results suggest that stronger static-scene completion does not suppress dynamic visual evidence in the observation branch.
\begin{figure}[t]
  \centering
  \includegraphics[width=\linewidth]{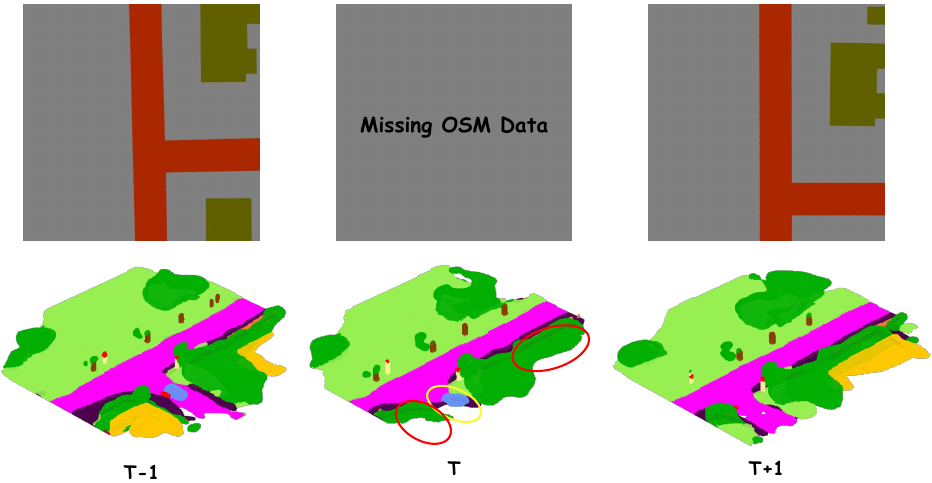}
  \caption{Failure case under missing OSM coverage. Highlighted regions show fragmented roads and unstable distant building boundaries when structured map guidance is unavailable.}
  \label{fig:missing_osm}
\end{figure}
\section{Additional Qualitative Results and Limitations}
\label{app:qualitative}

\begin{figure*}[t]
  \centering
  \includegraphics[width=\linewidth]{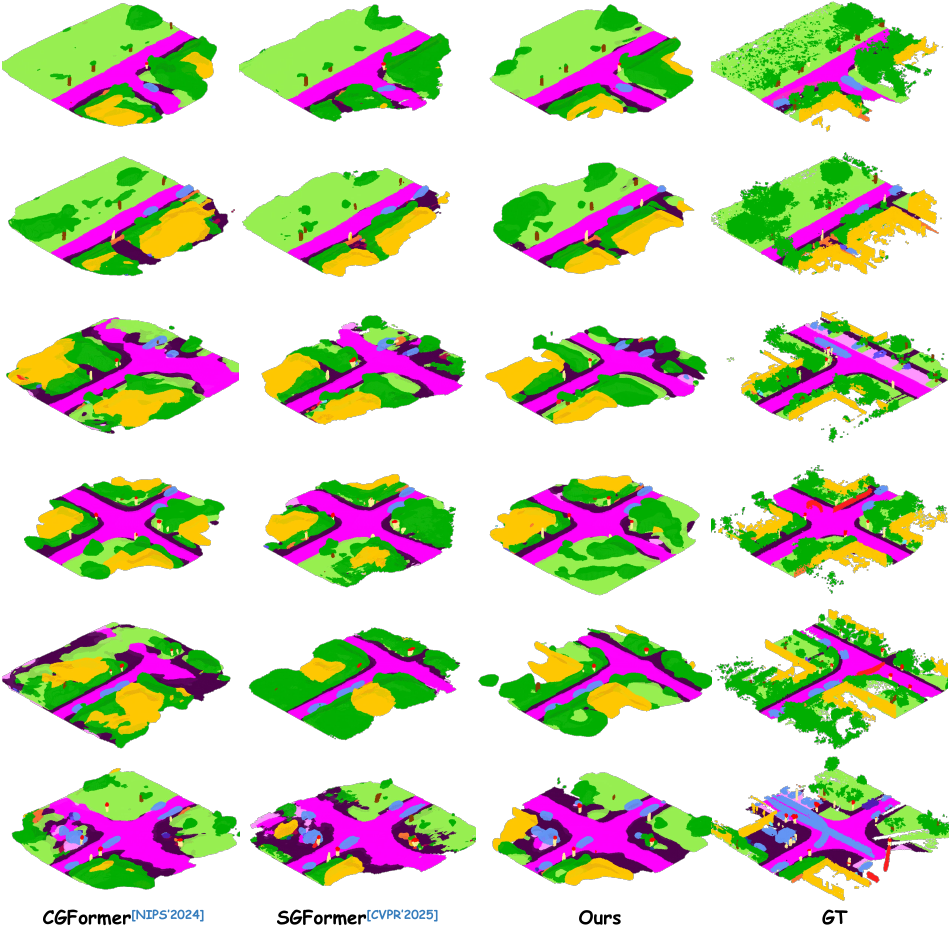}
  \caption{Additional qualitative comparisons on the SemanticKITTI validation set. GeoScene produces more continuous large-scale static structures, particularly beyond the directly observed camera region.}
  \label{fig:additional_qualitative}
\end{figure*}

Figure~\ref{fig:additional_qualitative} presents additional comparisons in challenging outdoor scenes. Compared with CGFormer and SGFormer, GeoScene generally reconstructs more continuous roads, clearer building extents, and fewer fragmented static regions. These differences are most evident outside the camera-observed region and for large, spatially persistent structures. Fine and dynamic objects remain primarily determined by onboard visual evidence.

\paragraph{Failure under missing map coverage.}
Figure~\ref{fig:missing_osm} shows a sequence with incomplete OSM coverage. Available neighbouring map primitives provide partial guidance for roads and buildings. When the OSM crop becomes empty, the model relies on onboard observations, and distant roads become more fragmented. Building boundaries also become less stable. Structural coherence improves when valid map information reappears. Thus, soft reliability weighting reduces dependence on unreliable priors but cannot replace global structure absent from both the map and onboard view.

\paragraph{Limitations.}
GeoScene depends on external geospatial resources whose coverage, temporal consistency, and image quality vary across regions. Satellite imagery may differ from the driving scene because of seasonal change, illumination, or construction. OSM may likewise be incomplete or outdated. The method primarily models static structure and does not explicitly represent temporary road closures or other dynamic map changes. The dual-weighting and voxel-refinement modules also add computation relative to purely image-based SSC models. Future work could incorporate timestamped geospatial snapshots, explicit map-change detection, and fallback training with missing or corrupted priors.

%% file: sec/tab/supp_tab.tex
\begin{table}[t]
  \centering
  \fontsize{9}{10.5}\selectfont
  \setlength{\tabcolsep}{1.2pt}
  \begin{tabular}{@{}cccccc@{}}
    \toprule
    \textbf{Cam.} & \textbf{Sat.} & \textbf{OSM} &
    \shortstack{\textbf{Geo-mIoU}\\\textbf{(\%)}} &
    \shortstack{\textbf{mIoU}\\\textbf{(\%)}} &
    \shortstack{\textbf{IoU}\\\textbf{(\%)}} \\
    \midrule
    \checkmark & & & 23.46 & 15.98 & 44.13 \\
    \checkmark & \checkmark & & 25.34 (+1.88) & 16.32 (+0.34) & 44.35 (+0.22) \\
    \checkmark & & \checkmark & 28.44 (+4.98) & 17.12 (+1.14) & 44.98 (+0.85) \\
    \checkmark & \checkmark & \checkmark & \textbf{32.56 (+9.10)} & \textbf{18.76 (+2.78)} & \textbf{46.58 (+2.45)} \\
    \bottomrule
  \end{tabular}
  \caption{Ablation of view transformation.}
  \label{tab:view}
\end{table}

\begin{table}[t]
  \centering
  \fontsize{9}{10.5}\selectfont
  \setlength{\tabcolsep}{5pt}
  \begin{tabular}{@{}cccc@{}}
    \toprule
    $\lambda_{\mathrm{bev}}$ & \textbf{Geo-mIoU (\%)} & \textbf{mIoU (\%)} & \textbf{IoU (\%)} \\
    \midrule
    0   & 30.96 & 17.45 & 45.44 \\
    0.5 & 30.36 & 17.78 & 45.96 \\
    1.0 & 30.98 & 18.15 & 45.87 \\
    1.5 & \textbf{32.56} & \textbf{18.76} & \textbf{46.58} \\
    2.0 & 31.43 & 18.09 & 46.06 \\
    2.5 & 30.98 & 17.87 & 45.38 \\
    \bottomrule
  \end{tabular}
  \caption{Sensitivity to the BEV loss weight $\lambda_{\mathrm{bev}}$.}
  \label{tab:loss}
\end{table}

\begin{table}[t]
  \centering
  \fontsize{9}{10.5}\selectfont
  \setlength{\tabcolsep}{3.5pt}
  \begin{tabular}{@{}ccccc@{}}
    \toprule
    $r$ & $o$ & \textbf{Geo-mIoU (\%)} & \textbf{mIoU (\%)} & \textbf{IoU (\%)} \\
    \midrule
    0.5 & 2.0 & 30.98 & 18.09 & 45.98 \\
    1.0 & 1.5 & 32.12 & 18.53 & 46.36 \\
    \textbf{1.5} & \textbf{1.0} & \textbf{32.56} & \textbf{18.76} & \textbf{46.58} \\
    3.0 & 0.5 & 32.32 & 18.23 & 46.23 \\
    \bottomrule
  \end{tabular}
  \caption{Sensitivity to the class-aware scaling coefficients $r$ and $o$ in Eq.~(4). The default setting is highlighted in bold.}
  \label{tab:ro_sensitivity}
\end{table}

\begin{table}[t]
  \centering
  \fontsize{9}{10.5}\selectfont
  \setlength{\tabcolsep}{5pt}
  \begin{tabular}{@{}cccc@{}}
    \toprule
    $\alpha$ & \textbf{Geo-mIoU (\%)} & \textbf{mIoU (\%)} & \textbf{IoU (\%)} \\
    \midrule
    0.3 & 31.33 & 17.99 & 46.03 \\
    0.5 & 31.64 & 18.25 & 46.15 \\
    \textbf{0.7} & \textbf{32.56} & \textbf{18.76} & \textbf{46.58} \\
    0.9 & 32.18 & 18.30 & 46.32 \\
    \bottomrule
  \end{tabular}
  \caption{Sensitivity to the fusion coefficient $\alpha$ in Eq.~(12). The default setting is highlighted in bold.}
  \label{tab:alpha_sensitivity}
\end{table}

\begin{table}[t]
  \centering
  \fontsize{9}{10.5}\selectfont
  \setlength{\tabcolsep}{4pt}
  \begin{tabular}{@{}lccc@{}}
    \toprule
    \textbf{Setting} & \textbf{Geo-mIoU (\%)} & \textbf{mIoU (\%)} & \textbf{IoU (\%)} \\
    \midrule
    Hard weights & 29.32 & 17.32 & 44.79 \\
    Soft weights & 30.21 & 17.54 & 44.85 \\
    DPWC (ours) & \textbf{32.56} & \textbf{18.76} & \textbf{46.58} \\
    \bottomrule
  \end{tabular}
  \caption{Ablation of soft reliability weighting strategies.}
  \label{tab:wei}
\end{table}

\begin{table}[t]
  \centering
  \fontsize{9}{10.5}\selectfont
  \setlength{\tabcolsep}{3.5pt}
  \begin{tabular}{@{}lccc@{}}
    \toprule
    \textbf{Setting} & \textbf{Geo-mIoU (\%)} & \textbf{mIoU (\%)} & \textbf{IoU (\%)} \\
    \midrule
    Concatenation & 29.76 & 17.12 & 44.79 \\
    Addition & 29.98 & 17.65 & 44.43 \\
    Attention & 30.45 & 17.87 & 45.32 \\
    WGVR (ours) & \textbf{32.56} & \textbf{18.76} & \textbf{46.58} \\
    \bottomrule
  \end{tabular}
  \caption{Ablation of voxel-level fusion and refinement strategies.}
  \label{tab:wgvr}
\end{table}

\begin{table}[t]
  \centering
  \fontsize{9}{10.5}\selectfont
  \setlength{\tabcolsep}{3.5pt}
  \begin{tabular}{@{}lccc@{}}
    \toprule
    \textbf{Setting} & \textbf{Geo-mIoU (\%)} & \textbf{mIoU (\%)} & \textbf{IoU (\%)} \\
    \midrule
    Highest-point & 31.02 & 17.54 & 46.01 \\
    Majority & 31.43 & 17.31 & 45.92 \\
    Geospatial-priority & \textbf{32.56} & \textbf{18.76} & \textbf{46.58} \\
    \bottomrule
  \end{tabular}
  \caption{Ablation of BEV target projection strategies.}
  \label{tab:pro}
\end{table}